\documentclass[pdflatex,sn-basic]{sn-jnl}

\usepackage{natbib}

\usepackage{multirow}

\usepackage{amsmath}

\usepackage{booktabs}

\usepackage[compatibility=false]{caption}
\usepackage{subcaption}

\usepackage{anyfontsize} 

\def\llamacpt{Llama-3.1-8B-Czech-Legal}
\def\modernbertcpt{ModernBERT-large-Czech-Legal}
\def\llamabase{Llama 3.1 8B Base}
\def\modernbertbase{ModernBERT Large}

\usepackage[table]{xcolor}
\colorlet{shadecolor}{yellow!25}

\begin{document} 

\title{Mining Legal Arguments to Study Judicial Formalism}

\author*[1,2]{\fnm{Tomáš} \sur{Koref}}\email{koref@c3s.uni-frankfurt.de}
\author[3,4]{\fnm{Lena} \sur{Held}}
\author[4]{\fnm{Mahammad} \sur{Namazov}}
\author[3]{\fnm{Yassine} \sur{Thlija}}
\author[4]{\fnm{Harun} \sur{Kumru}}
\author[4]{\fnm{Ivan} \sur{Habernal}}
\affil*[1]{\orgdiv{Center for Critical Computational Studies}, \orgname{Goethe University Frankfurt}, \orgaddress{\city{Frankfurt am Main}, \country{Germany}}}
\affil[2]{\orgdiv{Department for Legal Theory and Legal Doctrines, Faculty of Law}, \orgname{Charles University}, \orgaddress{\city{Prague}, \country{Czechia}}}
\affil[3]{\orgdiv{Trustworthy Human Language Technologies}, \orgname{Technical University of Darmstadt}, \orgaddress{\state{Germany}}}
\affil[4]{\orgdiv{Trustworthy Human Language Technologies}, \orgname{Research Center Trustworthy Data Science and Security of the University Alliance Ruhr \& Ruhr University Bochum}, \orgaddress{\country{Germany}}}

\abstract{Courts must justify their decisions, but systematically analyzing judicial reasoning at scale remains difficult. This study tests claims about formalistic judging in Central and Eastern Europe (CEE) by developing automated methods to detect and classify judicial reasoning in decisions of Czech Supreme Courts using state-of-the-art natural language processing methods. We create the MADON dataset of 272 decisions from two Czech Supreme Courts with expert annotations of 9,183 paragraphs with eight argument types and holistic formalism labels for supervised training and evaluation. Using a corpus of 300,511 Czech court decisions, we adapt transformer LLMs to Czech legal domain through continued pretraining and we experiment with methods to address dataset imbalance including asymmetric loss and class weighting. The best models can detect argumentative paragraphs (82.6\% Bal-F1), classify traditional types of legal argument (77.5\% Bal-F1), and classify decisions as formalistic/non-formalistic (83.8\% Bal-F1). Our three-stage pipeline combining ModernBERT, Llama 3.1, and traditional feature-based machine learning achieves promising results for decision classification while reducing computational costs and increasing explainability. Empirically, we challenge prevailing narratives about CEE formalism. We demonstrate that legal argument mining enables promising judicial philosophy classification and highlight its potential for other important tasks in computational legal studies. Our methodology can be used across jurisdictions, and our entire pipeline, datasets, guidelines, models, and source codes are available at \url{https://github.com/trusthlt/madon}.}

\keywords{Legal argument mining, judicial formalism, large language models, computational legal studies, Central and Eastern Europe}

\maketitle

\section{Introduction}

Courts are expected to justify their decisions.  How courts reason matters to scholars, practitioners, and citizens alike \citep{Walton_Macagno_Sartor_2021}. Yet systematically analyzing judicial reasoning at scale remains difficult. It is a challenging interdisciplinary task that inherits the key problems from legal NLP, legal theory, and empirical legal studies. Empirical work often depends on manual annotations, which are time-consuming and costly \citep{hall2008,gilardi_chatgpt_2023,braun_i_2023-1}. Legal concepts like “formalism" or traditional argument types remain disputed or abstract, hard to measure across large numbers of cases and foreign to the NLP community \citep{luders_classifying_2024-1,Habernal.et.al.2023.AILaw, schauer_1988}. While recent advances in NLP and legal argument mining promise to support quantitative computational analysis of legal reasoning, key challenges persist: clarifying complex legal concepts \citep{Habernal.et.al.2023.AILaw,luders_classifying_2024-1}, addressing scarce, imbalanced annotations with different quality \citep{braun_i_2023-1}, handling non-English legal texts, and ensuring explainable, reliable pipelines \citep{Chau_Livermore_2024}. In short, judicial reasoning matters, but studying it at scale remains hard.

We turn to Central and Eastern Europe (CEE), a region rarely examined empirically, yet widely alleged to be formalistic. Scholarship often portrays CEE judiciaries as rooted in late-communist legal culture and narrowly focused on text and procedure \citep{cserne_discourses_2020-1,kuhn_judiciary_2011-2,matczak_eu_2015,matczak_constitutions_2010-2,manko_weeds_2013,bystranowski_formalist_2022-1,Bencze_2021}. Scholars describe courts in the region as exhibiting “hyperpositivism" \citep{manko_weeds_2013}, “mechanical jurisprudence" \citep{Kosar2023}, and “vulgar" or “deeply rooted formalism" \citep{Suteu_2023,sipulova_kosar_purging_2024-1}. Formalism allegedly appears across the region, including Poland \citep{Kustra-Rogatka_2023, manko_weeds_2013}, Czechia \citep{kuhn_judiciary_2011-2, kosar_domestic_2020-1, Jakab_Dyevre_Itzcovich_2017}, Hungary \citep{matczak_constitutions_2010-2}, Romania \citep{Suteu_2023}, and Serbia \citep{Besirevic_2014}. But much of this is anecdotal. We still know little about how CEE courts in fact justify their decisions \citep{cserne_discourses_2020-1,cserne_judicial_2024}.

We ask three research questions. First, can state-of-the-art NLP methods detect argumentative paragraphs and classify them into eight traditional legal argument types in Czech judicial decisions? Second, can models distinguish between formalistic and non-formalistic reasoning styles at the decision level? And third, does an empirical analysis of legal reasoning in the Czech Supreme Court and Supreme Administrative Court support or challenge prevailing claims about formalism in Czech apex courts?

By addressing these research questions, we make five substantial contributions to the field. First, we create the MADON dataset, a collection of 272 Czech Supreme Court and Supreme Administrative Court decisions with 9,183 paragraphs annotated by legal experts with eight argument types and one holistic formalism label. Second, we publish detailed annotation guidelines grounded in established theories of legal argumentation and refined through over 200 pilot cases. Third, we compile and preprocess a corpus of 300,511 anonymized Czech court decisions for continued pretraining of LLMs. Fourth, we experiment with pretraining, fine-tuning and testing different configurations of state-of-the-art and traditional open models (Llama 3.1, ModernBERT, MLP) across three tasks: detecting arguments, classifying argument types, and predicting formalism. From these results, we propose a unified pipeline to classify decisions as formalistic/non-formalistic that integrates all tasks and model families, improving explainability, reducing computational costs, and providing the best performance. Finally, we use the MADON dataset for the first empirical analysis of the two Czech apex courts. In sum, we show how contemporary NLP research helps to investigate judicial philosophies and traditional jurisprudential questions; we provide novel datasets, pipelines, and models for such investigations.

\section{Related work}
\subsection{Jurisprudential debate on formalism}

The following simple example illustrates formalistic judging. Suppose a statute fixes a 15-day deadline to file an appeal. If a party files on day 16 because the counsel was hospitalized, a formalistic judge would dismiss the appeal as inadmissible, since the statute contains no exception. A non-formalistic judge would accept the appeal, invoking access-to-justice principles, the purpose of the rule, and the unfairness of harsh consequences under the individual circumstances.

Formalism concerns how lawyers, typically judges, might approach legal questions. It is an approach to law and judicial adjudication that prioritizes the binding authority of legal norms and their wording. A formalist sees law as distinct from politics and limits the influence of non-legal norms and extra-legal considerations in answering legal questions, such as social, economic, political, or moral factors \citep{weber_1922,schauer_1988,Weinrib1988}.

The jurisprudential debate on legal formalism is longstanding and widespread. Some authors argue that the term has lost all meaning, while others have sought to rediscover and clarify it \citep{schauer_1988,Nachbar2020}. The debate spans sociology of law \citep{weber_1922}, jurisprudence and legal argumentation \citep{schauer_1988}, Law and Economics \citep{Djankov_2002} and critical legal studies \citep{Kennedy2004Disenchantment}. Despite the variety of perspectives on formalism, prominent scholars from these different traditions usually discuss the question already noted at the outset: to what extent are the judges bound by existing legal rules and their wording?  

Formalism can be approached analytically, normatively, or descriptively. The analytical approach asks, e.g., what formalism is and whether it is possible (e.g., can a rule restrict a judge given the very same legal system already contains open-textured principles). The normative approach asks whether formalism is a proper way to deal with legal questions---is it right that judges stick to statutory wording rather than purpose? Lastly, the descriptive approach investigates whether judges in fact decide cases in a formalistic manner (e.g., would an average judge indeed stick to the letter of the law and disregard general values?). For example, \citet{schauer_1988} combines analytical and normative approach to clarify what formalism is and why it is desirable. An illustrative example of a descriptive approach would be that of \citet{stiglitz_h_historical_2024} who convincingly investigate how the formalism of US Supreme Court changed throughout history or whether formalistic judges stay formalist \citep{Stiglitz2026}. The debate on CEE combines the descriptive and normative aspects by claiming that formalism is omnipresent and undesirable. We introduce this debate in the next section.

\subsection{Legal formalism in Central and Eastern Europe (CEE)}
\label{sec:tale.of.two.courts}

Descriptively, formalism is widely portrayed as the dominant judicial style in the CEE region \citep{bystranowski_formalist_2022-1,dixon_2023,Bencze_2021,kuhn_judiciary_2011-2,Kustra-Rogatka_2023,sipulova_kosar_purging_2024-1}. Some authors even suggest that the socialist past and formalist present might justify classifying Central European legal systems into a distinct “Central European Legal Family" \citep{MankoCercelSulikowski2016, Manko2013_Legal_tradition}. In short, existing literature sees CEE judiciaries as distinct and formalistic.

From the normative perspective, formalism is heavily criticized in the CEE region. Scholars argue that it impedes rule of law \citep{matczak_2018} and the enforcement of EU law \citep{matczak_constitutions_2010-2,kuhn_worlds_2004}. According to \citet{kuhn_judiciary_2011-2}, the formalism shows that the transition to liberal democracy was weak or incomplete, given the alleged communist-era roots of formalism. These criticisms are echoed in practice. The Czech Constitutional Court has repeatedly criticized the rest of the judiciary for excessive formalism and struck excessive formalism down as a violation of fundamental rights \citep{constitutionalcourt2010}.\footnote{The court held: “The Constitutional Court has repeatedly demonstrated in its consistent case law that it does not tolerate, on the part of public authorities, and in particular the ordinary courts, a formalistic approach that employs sophisticated reasoning to enforce an evident injustice. (...) Excessive formalism in the interpretation of legal norms leading to an extremely unjust conclusion constitutes a violation of fundamental rights."} Alleged formalism has also been raised in judicial appointments to constitutional courts as a reason to nominate more liberal judges \citep{CzechSenate2002}. Beyond the scholarship and courts, politicians have critiqued formalism too: the Slovak President recently blamed judicial formalism for public frustration in post-communist Slovakia and Czechia \citep{CT24_2024_Caputova}. Formalism has become central to debates about the region's communist past and democratic future. Key actors view formalism as a toxic post-communist legacy undermining judicial credibility and the rule of law in the new democratic state.

The formalism debate resonates particularly strongly in Czechia, where it concerns the country's two supreme courts: the Supreme Court (SC), handling civil and criminal matters, and the Supreme Administrative Court (SAC), deciding administrative matters. The SC is characterized as an unreformed formalistic institution inherited from the communist era \citep{kuhn_worlds_2004,kuhn_application_2005,kuhn_judiciary_2011-2,kuhn_quality_2018,matczak_constitutions_2010-2,matczak_eu_2015,kosar_domestic_2020-1,sipulova_kosar_purging_2024-1,Kadlec2016,stehlik_unijni_2014}. By contrast, the SAC, created in 2003 and staffed by a new generation of jurists, is described as “discursive" and more open in its reasoning \citep{matczak_eu_2015}. Allegedly, the SC is a formalistic post-communist court, while the SAC is not.

Despite the omnipresent debates, little is known about how CEE judiciaries actually decide cases, including the SC and SAC. Some of the recent scholarship rightly notes that the formalism narrative lacks clear definitions and remains mostly anecdotal \citep{cserne_judicial_2024,cserne_discourses_2020-1,komarek_2015,manko_weeds_2013}. The evidence is missing.

This paper aims to close the existing research gap by a pioneering empirical research in this direction. We next summarize current approaches to empirical analysis of formalism and how we build on them.

\subsection{Existing approaches to empirical analysis of formalism} 
The literature offers three approaches to the empirical analysis of formalism.

The first approach studies the types of arguments courts employ. It quantifies the frequency of various argument types (e.g., linguistic, systemic, teleological) that are then classified as either formalistic (e.g., linguistic arguments) or non-formalistic (e.g., teleological arguments). This approach is not new; legal theory has investigated argument types for centuries, and typically US scholars like \citet{choi_2020} or \citet{Krishnakumar2020,krishnakumar_textualism_2024} have been empirically investigating argument types to assess prevailing interpretative theories. Fewer studies focus on argument types to measure formalism specifically. Most of these concern CEE judiciaries \citep{matczak_constitutions_2010-2,matczak_eu_2015,bystranowski_formalist_2022-1}. The underlying assumption of this approach is that the more formalistic arguments a court uses, the more formalistic the court is.

The second approach \citep{Stiglitz2026, stiglitz_h_historical_2024,thalken_modeling_2023} directly labels paragraphs as formalistic or non-formalistic without analyzing specific argument types. This simplification enabled them not only to manually code decisions but also to provide large-scale analysis of US Supreme Court using NLP. This approach assumes that the higher the number of formalistic paragraphs appearing in courts' decisions, the more formalistic the court is.

The third approach captures different dimensions of formalism by examining various parameters simultaneously, including the court's rhetorical style and procedural choices alongside argumentation patterns. This approach offers a holistic perspective on how formalism manifests across different dimensions of judicial decision-making and has been applied mainly in an Israeli context \citep{alberstein_sarat_measuring_2012, alberstein_whats_2020}.

Current approaches face several limitations. Most rely on small, manually coded samples, cover only narrow slices of case law, and have reproducibility issues as they do not publish annotation manuals or report inter-annotator agreement \citep{matczak_eu_2015,bystranowski_formalist_2022-1}. None use argument mining to measure formalism directly. Each approach also has specific weaknesses: simply counting formalistic argument or paragraphs can mislead when a single decisive principle outweighs numerous peripheral textual references; paragraph-level labeling downplays argument types, the standard lens in legal theory \citep{Walton_Macagno_Sartor_2021,alexy_argument_2010} and provides less informative data on reasoning practices in detail; and the multiparameter approach, though comprehensive, brings significant complexity to annotation.

We combine the best of all three approaches. We analyze argument types with detailed annotation guidelines and reported inter-annotator agreement; we combine fine-grained argument mining across eight types with a binary holistic formalism label with a NLP pipeline that reduces computational requirements and increases explainability; and we condense multiple dimensions of formalism into a single binary label while employing NLP to predict it. Our approach thus enables richer measurement of judicial reasoning by making use of recent advances in legal argument mining, which we introduce next.

\subsection{Legal argument mining}
Argument mining aims to automatically identify and extract arguments \citep{lawrence_argument_2019}. Legal argument mining (LAM) focuses on the legal domain and legal arguments \citep{Habernal.et.al.2023.AILaw,zhang2022, MochalesMoens2011}. Recent research has applied legal argument mining to classify the traditional types of legal arguments \citep{Habernal.et.al.2023.AILaw,luders_classifying_2024-1}. This emerging field has been growing in recent years and increasingly intersects with empirical legal research. For current advances, we refer to recent LAM papers \citep{Habernal.et.al.2023.AILaw,luders_classifying_2024-1} as well as a systematic literature reviews of the field and related workshops \citep{zhang2022, liepina2025legal, amelr2025preface}.

Several related studies, while not framed as LAM, also classify legal arguments or interpretive theories in judgments using machine learning and NLP. Examples include \citet{Peters2024} on US state courts,  \citet{DugacAltwicker2025} on the ECtHR, \citet{stiglitz_h_historical_2024} on the entire U.S. Supreme Court corpus (1870–2023), \citet{MunozSoro_etal_2024} on Spanish custody judgments or work on U.S. tax statutory interpretation using traditional ML techniques \citep{choi_2020}.

To our knowledge, however, no study has used LLMs or other ML methods to examine interpretive theories of CEE courts.

\section{Operationalizing and annotating formalism}
\label{sec:operationalizing}

To train supervised classifiers for detecting formalism, we operationalize an abstract legal concept into concrete, annotatable labels through “core tenets of CEE formalism". 

\subsection{Five core tenets of CEE formalism}

Current legal scholarship on formalism in CEE converges on five distinctive criteria of formalistic judicial decision-making. Formalistic courts in the CEE would
\begin{enumerate}
\item rely heavily on a limited set of arguments derived from statutory text \citep{matczak_constitutions_2010-2,matczak_eu_2015,bystranowski_formalist_2022-1,Jakab_Dyevre_Itzcovich_2017,cserne_discourses_2020-1,kuhn_judiciary_2011-2},
\item focus on the most locally applicable rule, not broader principles \citep{matczak_constitutions_2010-2,matczak_eu_2015,bystranowski_formalist_2022-1,manko_weeds_2013,cserne_discourses_2020-1,Bencze_2021,kuhn_judiciary_2011-2},
\item usually exclude ``external standards'' in reasoning, like efficiency, justice, moral and political reasoning \citep{kuhn_worlds_2004,matczak_constitutions_2010-2,bystranowski_formalist_2022-1,MalolepszyGluchowski2021}, {teleological interpretation } \citep{kuhn_worlds_2004,kuhn_judiciary_2011-2,Jakab_Dyevre_Itzcovich_2017,MalolepszyGluchowski2021} {or even historical interpretation} \citep{matczak_constitutions_2010-2,matczak_eu_2015},
\item dismiss cases on formal grounds to avoid analyzing them on the merits \citep{manko_weeds_2013, bystranowski_formalist_2022-1, uzelac_survival_2010}, and 
\item provide scarce reasoning for their decisions  \citep{kuhn_worlds_2004,Suteu_2023,kuhn_judiciary_2011-2}.
\end{enumerate}
We call these five criteria \emph{the core tenets of CEE formalism}. 

Most of the tenets are not CEE specific. Although we extracted the core tenets from a thorough review of key CEE scholarship as we aim to empirically test existing claims about the CEE region, the main tenets appear shared beyond CEE. Similar criteria appear across very different regions and periods--from Weber's sociology of law a century ago \citep{weber_1922} and Schauer's influential jurisprudential article in the 1980s \citep{schauer_1988} to more recent computational work by \citet{stiglitz_h_historical_2024}. In these works, reliance on the text of legal norms, focus on the most locally applicable rule rather than general values or principles, and avoidance of political, social, and economic reasoning often appear as important aspects of formalism.

\subsection{Argument types} \label{subsec:measure_formalism}

The three core tenets of CEE formalism all concern legal arguments, but they are too abstract for direct annotation of real-world decisions. For example, the first tenet (heavy reliance on text-based arguments) requires specifying what counts as text-based argument and how to identify it in the judgments. Without such specification, empirical analysis remains impossible.

To address this challenge, we developed annotation guidelines iteratively over several months. We started from leading argumentation taxonomies \citep{alexy_argument_2010,maccormick_interpreting_2016,Walton_Macagno_Sartor_2021} and German and Czech legal scholarship \citep{Mollers2020LegalMethods,Wintr2019Metody,LarenzCanaris_Methodenlehre_1995}, which we refined on more than 200 pilot decisions before annotating the MADON dataset. The guidelines identify eight argument types, classified as formalistic or non-formalistic based on the core tenets of CEE formalism:

\begin{table}[h]
\centering
\small
\begin{tabular}{l p{4cm} @{\hspace{1cm}} l p{4cm}} 
    \toprule
    \multicolumn{2}{c}{\textbf{FORMALISTIC}} & \multicolumn{2}{c}{\textbf{NON-FORMALISTIC}} \\
    \cmidrule(r{1cm}){1-2} \cmidrule{3-4}
    \textbf{LIN} & Linguistic Interpretation & \textbf{HI} & Historical Interpretation \\
    \textbf{SI}  & Systemic Interpretation & \textbf{PL} & Principles of Law/Values \\
    \textbf{CL}  & Case Law                & \textbf{TI} & Teleological Interpretation \\
    \textbf{D}   & Doctrine                & \textbf{PC} & Practical Consequences \\
    \bottomrule
\end{tabular}
\end{table}
    
The first three core tenets map directly to four of our argument types. Heavy reliance on text (Tenet 1) corresponds to Linguistic Interpretation; narrow focus on rules rather than principles (Tenet 2) corresponds to low frequency of Principles of Law and Values; and exclusion of external standards (Tenet 3) corresponds to low frequency of Teleological Interpretation and Practical Consequences. 

We classify Systemic Interpretation, Case Law, and Doctrine as formalistic, and Historical Interpretation as non-formalistic, consistent with previous studies \citep{matczak_constitutions_2010-2, matczak_eu_2015}. Our underlying reasoning is that Systemic Interpretation reflects the internal hierarchy of legal norms; Case Law relies on the authority of prior decisions and rules extracted from them; and Doctrine provides authoritative guidance, typically through commentaries that systematize case law.

Historical Interpretation is more complicated. In Germany, it could even be seen as formalistic due to its link to the “subjective theory" of interpretation, which is often contrasted with the more open “objective theory" focusing on the purpose of the law \citep{Mollers2020LegalMethods}. In contrast, US textualists reject legislative history altogether \citep{scalia1997matter} and US empirical studies accordingly connect legislative history with non-formalistic reasoning \citep{stiglitz_h_historical_2024}. Similarly, we classify Historical Interpretation as non-formalistic. In this decision, we mainly reflect the CEE scholarship, which often considers legislative history as distinct from the more formalistic methods related to the text of the law \citep{matczak_constitutions_2010-2,Wintr2019Metody}. In practical terms, this classification has minimal impact on our results, as Historical Interpretation is the least common argument type in our dataset amounting to only 3\% of all arguments.

For each argument type, our guidelines included a structured description that connects the theoretical concept to real‑world decisions. For most types, we (1) give a brief description, (2) list its subtypes, (3) note “trigger” phrases that usually signal its presence in a decision, (4) provide typical and borderline examples, (5) add comments and annotator tips and (6) include annotation chart consistent with \citet{thalken_modeling_2023}  to guide the annotation process. The creation of these guidelines  itself represents an important contribution to the field.\footnote{Appendix A contains the summary of the guidelines. Full version of the guidelines is accessible at \url{https://github.com/trusthlt/madon}.} 

We then applied the guidelines to annotate the below described MADON dataset of 272 decisions from Supreme Court and Supreme Administrative Court of Czechia. We used a set of cases distinct from the pilot decisions to support reliability of the annotation.

Court decisions are structured into paragraphs, and we followed this structure by annotating at the paragraph level rather than sentence level. Our interest was in whether a given argument type is present in a paragraph, not in its exact boundaries. Each argument type was counted once per paragraph; each paragraph could thus include between zero and eight arguments. We annotated only the reasoning ultimately adopted and we excluded arguments the court explicitly rejected similarly to \citet{Krishnakumar2020}.

\subsection{Holistically labelling judgments as formalistic}

Although arguments are important, we found we should not reduce formalism to simple quantification of arguments. In our pilot studies, some decisions dominated by formalistic arguments turned out non-formalistic once assessed holistically. Three reasons explain why counts alone can mislead. First, arguments differ in weight (some arguments are more important for the outcome while others are peripheral); context matters (a decision criticizing lower courts for excessive formalism may still cite mostly statutory text); and the last two tenets, i.e., dismissal on formal grounds and insufficient justification, are not capturable by simply counting arguments.

We therefore annotated each decision with a binary holistic label (formalistic/non-formalistic). The expert decision is based on a complex assessment of the entire case. Concretely, the annotators assessed each decision against all five core tenets. The annotators also considered the frequency, weight and centrality of individual arguments---for instance, whether a single decisive reference to a legal principle determined the result and outweighed numerous peripheral citations to statutory text, as well as other contextual and factual matters, like whether the court itself critiqued lower courts for excessive formalism. The label thus reflects an expert judgment about the decision as a whole. Our annotation guidelines include the instructions and annotation chart used for annotation.\footnote{\url{https://github.com/trusthlt/madon}}

\subsection{Annotation process}

The systematic annotation of Czech Supreme Courts' decisions was conducted between July and September 2024, building on a pilot study from early 2024 that analyzed 160 decisions and helped refine the methodology. We employed four law students from the Faculty of Law, Charles University, as research assistants, who were led by a PhD student in legal argumentation. All of them had a background in legal theory and argumentation.

The process had four phases: Introduction (Week 1), Training (Weeks 2–6, 60+ decisions), Coding (Weeks 7–12, 272 decisions with regular reviews), and Finalization (Week 13). The team met on a weekly basis to discuss the annotations in the Introduction and Training phase. We used the INCEpTION \citep{tubiblio106270} interface for accessible annotation.  We followed methodological guidance for the legal domain and drew mainly from content analysis literature \citep{hall2008, Ovádek_Schroeder_Zglinski_2024} and recommended practices by \citet{braun_i_2023-1} for annotating legal NLP datasets. This approach involved several hundred hours of training and guideline development, with at least two independent annotators coding each decision in the MADON dataset. We measured inter-annotator agreement and report on the annotation process. In total, the annotation took 1,000+ hours, covered more than 450 decisions including pilot studies (272 after pilot), and produced the MADON dataset with 272 decisions consisting of 9,183 annotated paragraphs, including 1,913 legal arguments in total and 1,237 paragraphs with at least one argument.

\subsubsection{Inter-annotator agreement and dataset curation}

 To measure reliability, we calculated two different inter-annotator agreement metrics corresponding to our two types of labels: Cohen's kappa for the holistic label and Krippendorff's unitized alpha for argument types. 

Our holistic assessment achieved a Cohen’s kappa coefficient of 0.65 on the overall category formalistic/non-formalistic, indicating ``substantial agreement'' among annotators despite the complexity of the assessment \citep{Landis_Koch_1977}. This promising result yields two significant findings for future empirical research. First, research assistants can be effectively trained to achieve sufficient inter-coder agreement on complex holistic assessments of formalistic/non-formalistic reasoning. Second, we found a strong correlation between decisions with higher frequencies of non-formalistic arguments and holistic non-formalistic assessments; we use this finding to build original NLP architecture that predicts the holistic label based on the arguments.

\begin{table}
\caption{Inter-annotator agreement in Krippendorff's $\alpha_u$  for each argument type}
\label{tab:parts_compare}
\centering
\begin{tabular}{p{5cm}rr}
\toprule
\textbf{Argument} & \textbf{Part I} & \textbf{Part II} \\
\midrule
Linguistic Interpretation (LIN) & 0.54 & 0.28 \\
Systemic Interpretation (SI) & 0.35 & 0.42 \\
Case Law (CL) & 0.95 & 0.94 \\
Doctrine (D) & 0.94 & 0.90 \\
Historical Interpretation (HI) & 0.68 & 0.80 \\
Principles of Law and Values (PL) & 0.76 & 0.65 \\
Teleological Interpretation (TI) & 0.63 & 0.65 \\
Practical Consequences (PC) & 0.20 & 0.21 \\
\bottomrule
\end{tabular}
\end{table}

For argument type annotation, we calculated Krippendorff's unitized alpha across two parts of the annotation process (in total 272 decisions), see Table~\ref{tab:parts_compare}.  According to Krippendorff, sufficient agreement requires $\alpha \geq 0.800$, while data with agreement $0.800 > \alpha \geq 0.667$ can be used to draw tentative conclusions \citep{Krippendorff_2019}. Half of our categories meet the 0.667 threshold, while three categories (Linguistic Interpretation, Systemic Interpretation, and Practical Consequences) demonstrate significantly lower reliability.

This pattern reflects the inherent complexity of legal document annotation. Annotation of legal documents is generally considered challenging, and inter-annotator agreement is often lower \citep{braun_i_2023-1,vanDijck2024}. Existing literature identifies three main sources of annotator disagreement: annotators themselves, annotation guidelines, and the concepts of interest \citep{Ovádek_Schroeder_Zglinski_2024}. 

We believe the problem with the three categories (Linguistic Interpretation, Systemic Interpretation, and Practical Consequences) lies primarily with the concepts. Notably, most disagreements were reasonable, i.e., good reasons existed for either annotation, and clear mistakes or deviations from guidelines were rare.

For the three low-agreement argument types, we were unable to resolve the ambiguity despite extensive efforts in our pilot studies. We tried repeated revision rounds \citep{vanDijck2024}, better definitions, label refinement (e.g., merging constitutional conforming interpretation into systemic interpretation), additional examples, and flowcharts \citep{thalken_modeling_2023}. These techniques helped but produced only modest improvements for the three argument types. The disagreements appear to reflect a deeper problem, namely that legal theory remains abstract and imprecise when applied to actual decisions. Most real-world cases are not, by definition, the prototypical examples from the textbooks, and the argument types themselves prove contentious and unstable in practice, even though they are widely used and broadly understood at the theoretical level. Adding more examples or decision rules did not resolve this, because the problem seems to lie with the concepts, not the guidelines.

We retain the three low-agreement categories despite the lower agreement. We do so for two main reasons. First, excluding labels due to low inter-annotator agreement is not standard practice in legal NLP; many studies either do not employ multiple annotators or do not report agreement at all \citep{braun_i_2023-1}. Second reason is appropriate curation. We followed the recommended curation strategy; every disagreement was resolved by an arbiter, a co-author with legal background specializing in legal argumentation. The arbiter evaluated each case against the annotation guidelines and existing theories of legal argumentation before assigning one of the labels proposed by the original annotators. Every label in the dataset was thus approved by at least two people, with contested labels decided by a more senior domain expert. The annotator disagreement thus served a twofold goal; it first helped refine the guidelines during pilot studies, and later flagged for senior review those ambiguous cases that include something potentially relevant for our project. While lower agreement introduces some necessary noise, this curation strategy reduces its negative impact on the validity and consistency of the dataset. We follow the best practices and publish not only the curated data, but also all labels from all annotators.

For the holistic label, we employed a combination of arbiter review and “forced agreement" \cite{braun_i_2023-1}. This meant that annotators were required to discuss their disagreements and propose potential solutions, accompanied by justifications for their choices. The arbiter then reviewed both the suggested solutions and their justifications; when annotators agreed on how to solve initial disagreement but the arbiter disagreed with their consensus, the arbiter assigned the final label (i.e., the initial label of one of the annotators).

\section{MADON dataset}

The MADON\footnote{MADON stands for Mining Arguments to Debunk Old Narratives} is the annotated dataset resulting from the process described in Section~\ref{sec:operationalizing} and comprises 272 legal decisions from the Supreme Court and the Supreme Administrative Court, the only two Supreme Courts.\footnote{Like Germany, Austria, Poland and many other countries in Europe, Czechia has a separate Constitutional Court “above" the Supreme Courts. It hears mainly constitutional complaints about alleged violations of fundamental rights. The Constitutional Court is significantly distinct from the Supreme Court and the Supreme Administrative Court and is not included in our dataset.}

\subsection{Stratified sampling of documents for expert annotation}

The MADON dataset spans  the period from 1997 to 2024, with a focus on decisions made between 2003 and 2023. We employed stratified sampling to ensure a representative sample. The stratified sample reflects time periods, court agendas (civil, criminal, and administrative), and types of cases (procedural and on the merits).
This method ensured proportional representation across the Supreme Court's civil branch (122 cases), criminal branch (58 cases), and the Supreme Administrative Court (90 cases). Within these strata, we drew random samples.

\subsection{MADON statistics and first empirical findings} \label{subsec: madon_stats}

\begin{figure}
  \centering
  \begin{subfigure}[t]{0.48\linewidth}
    \centering
    \includegraphics[width=\linewidth]{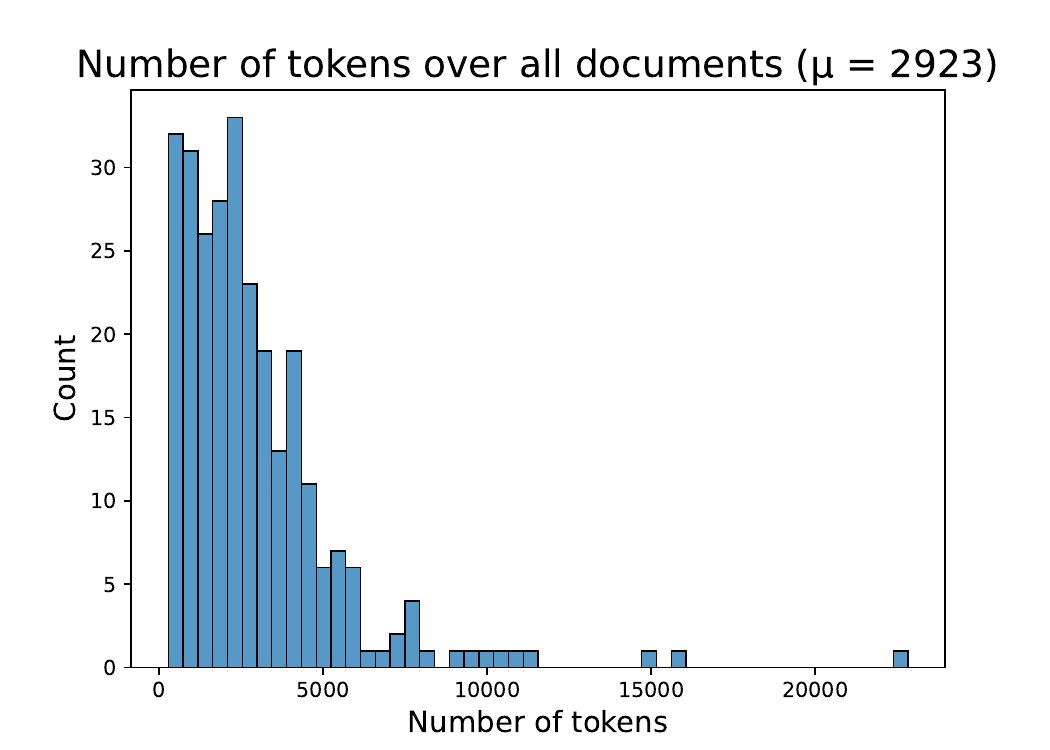}
    \subcaption{Document length distribution}
    \label{fig:hist_tokens}
  \end{subfigure}\hfill
  \begin{subfigure}[t]{0.48\linewidth}
    \centering
    \includegraphics[width=\linewidth]{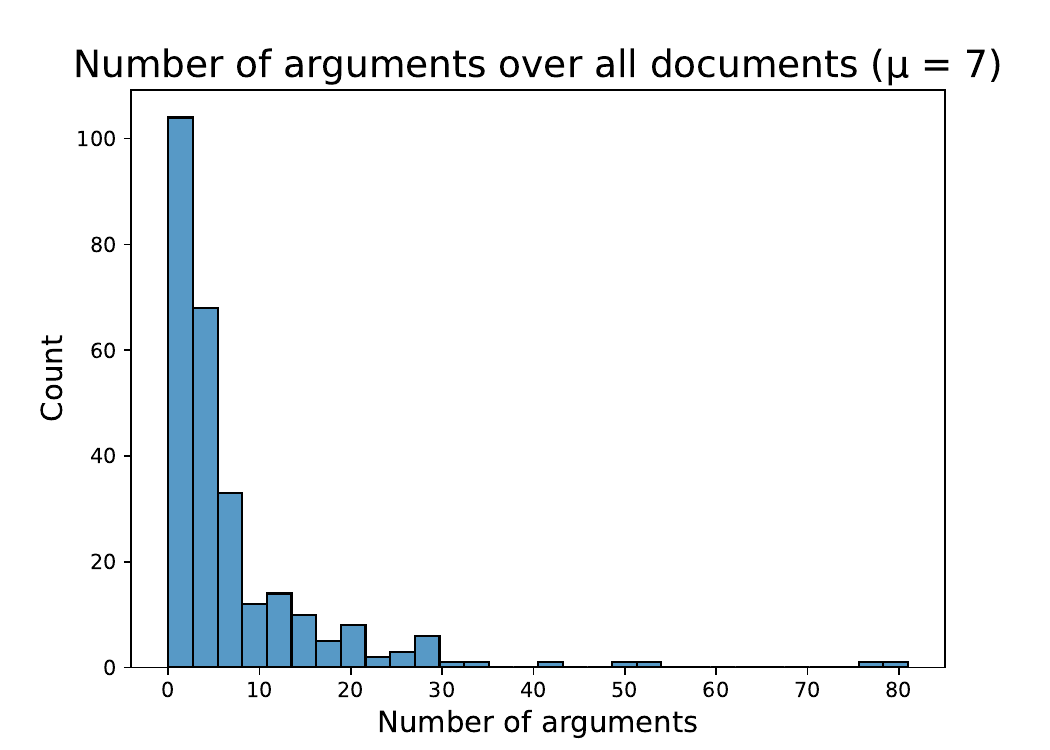}
    \subcaption{Number of arguments distribution}
    \label{fig:hist_args}
  \end{subfigure}
  \caption{\label{fig:tokens_args_hists} Documents in the MADON dataset are (a) long on average but rarely exceeding 6,000 tokens, and contain 7 arguments on average (b) with some outliers with almost 80 arguments.}
\end{figure}

The MADON dataset contains 9,183 paragraphs with 1,913 annotated arguments. 765 paragraphs contain one argument, 472 paragraphs contain 2 or more arguments. 7,946 paragraphs (87\%) contain no arguments.
Individual decisions are highly heterogeneous. The shortest decision runs to 275 tokens, the longest to 22,831, with a mean of 2,923 tokens. The average number of arguments is 7, with a maximum of 81 arguments. Notably, 43 documents contain no arguments at all; see Figure~\ref{fig:tokens_args_hists} for histogram plots.
This heterogeneity is caused by the fact that we intentionally represented all types of decisions, including procedural dismissals that are often short.

\begin{figure}
  \centering
  \includegraphics[width=.55\linewidth]{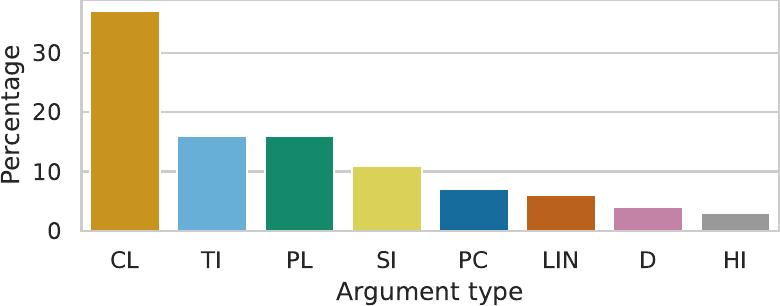}
  \caption{\label{fig.arg.type.dist} Without distinguishing between the two courts, the most prevalent argument type for the entire period 1997--2024 is Case Law (CL).}
\label{types}
\end{figure}

The distribution of arguments reveals surprising patterns (see Figure~\ref{fig.arg.type.dist}). Linguistic interpretation (LIN), despite its theoretical centrality to formalism, accounts for only 5.7\% of all arguments across both courts in 1997–2024 (108 occurrences). By contrast, two non-formalistic arguments, namely legal principles and values (PL) and teleological interpretation (TI), together make up 32.3\%. Most striking is the dominance of case law (CL), which at 37.4\% (707 occurrences) is the single most common argument type. These patterns challenge prevailing stereotypes: the SC is not text-bound, nor does it differ markedly from the SAC in argumentation profile. More broadly, Czech courts' scarce use of explicit text-based reasoning and frequent reliance on purposive and value-based arguments diverges  from formalism as understood in both CEE and U.S. contexts, where it is closely tied to textualism. We thus empirically challenge the claim that Czech courts are text-bound and had been precedent-averse \citep{manko_weeds_2013,matczak_eu_2015,Spamann_2024}.

While the argumentation profiles are similar at SC and SAC, the distribution of holistic formalism/non-formalism labels differs across courts. The SC had 116 formalistic (64\%) and 66 non-formalistic (36\%) decisions for a total of 182 cases. The SAC had 45 formalistic (50\%) and 45 non-formalistic (50\%) decisions for a total of 90 cases. Non-formalistic decisions are more frequent especially in the later period (2014-2024). This indicates that the alleged differences between the SC and SAC, i.e., the claim about formalistic SC and non-formalistic SAC, became closer to reality  in the later period, where SAC moved toward more non-formalistic decisionmaking. These findings are described in more detail in our complementary work \citep{koref2025}. The findings are based on the MADON dataset and await verification through our downstream large-scale analysis. In this paper, we use the MADON dataset primarily for the NLP experiments described in the following section.

\begin{table}
\caption{Distribution of argument types across paragraphs and documents}
\label{tab:arg_distribution}
\centering
\small
\begin{tabular}{llrr}
\toprule
\textbf{Type} & \textbf{Argument} & \textbf{Paragraphs} & \textbf{Documents} \\
\midrule
\multirow{4}{*}{Formalistic}
 & Linguistic Interpretation (LIN) & 108 & 63 \\
 & Systemic Interpretation (SI)    & 208 & 78 \\
 & Case Law (CL)                   & 707 & 181 \\
 & Doctrine (D)                    & 68  & 39 \\
\midrule
\multirow{4}{*}{Non-formalistic}
 & Historical Interpretation (HI)  & 58  & 22 \\
 & Principles of Law and Values (PL) & 311 & 117 \\
 & Teleological Interpretation (TI) & 315 & 124 \\
 & Practical Consequences (PC)      & 138 & 68 \\
\bottomrule
\end{tabular}
\end{table}


\section{Experiments}

\subsection{Experiment setup and tasks}
We decompose the problem into three tasks of increasing complexity, following standard practice in argument mining~\citep{eger-etal-2017-neural}:
\begin{itemize}
\item \textbf{Task 1} (argument presence detection) classifies paragraphs as \textit{argumentative} or \textit{non-argumentative}.
\item \textbf{Task 2} (argument type classification) assigns up to eight argument type labels per paragraph in a multilabel setting.
\item \textbf{Task 3} (holistic formalism classification) classifies entire decisions as \textit{formalistic} or \textit{non-formalistic}.
\end{itemize}

\subsection{Data for NLP experiments}

For the experiments, we split the MADON dataset into train (70\%), validation (20\%), and test (10\%) sets. The split was stratified to proportionally represent holistic labels and issuing court throughout the subsets, as summarized in Table~\ref{tab:madon_compact}.

In addition, we compiled a new corpus for continued pretraining of our models. We collected and preprocessed 300,511 decisions from the Czech Supreme Court (152,925) and all administrative courts (147,586), comprising the Supreme Administrative Court (74,937) and eight Regional Administrative Courts (72,631). We make the corpus available to the AI\&Law research community.\footnote{\url{https://github.com/trusthlt/madon}} The decisions were collected from the official anonymized databases of the respective courts. Our corpus complements existing Czech legal datasets, namely \citep{novotna2019czechcourtdecisionscorpus} and \citep{Paulik_2025}. Compared to \citet{novotna2019czechcourtdecisionscorpus}, we provide regional administrative court decisions, richer metadata, scraping scripts, and, most importantly, coverage of the six year period after their cut-off date, i.e., 09/2018-09/2024. The dataset from \citet{Paulik_2025} focuses on the Constitutional Court, a different institution not covered here.

\begin{table}
\caption {MADON dataset splits with label and court distributions}
\label{tab:madon_compact}
\centering
\begin{tabular}{@{}lrrrr@{}}
\toprule
 & \textbf{Train} & \textbf{Validation} & \textbf{Test} & \textbf{Total} \\
\midrule
\multicolumn{5}{l}{\textbf{Holistic Labels}} \\
\quad Formalistic         & 112 (59.3\%) & 32 (59.3\%) & 17 (58.6\%) & 161 (59.2\%) \\
\quad Non-Formalistic     &  77 (40.7\%) & 22 (40.7\%) & 12 (41.4\%) & 111 (40.8\%) \\
\addlinespace[0.3em]
\multicolumn{5}{l}{\textbf{Court Distribution}} \\
\quad Supreme Court           & 127 (67.2\%) & 36 (66.7\%) & 19 (65.5\%) & 182 (66.9\%) \\
\quad Supreme Admin. Court    &  62 (32.8\%) & 18 (33.3\%) & 10 (34.5\%) &  90 (33.1\%) \\
\midrule
\textbf{Total} & \textbf{189} & \textbf{54} & \textbf{29} & \textbf{272} \\
\bottomrule
\end{tabular}
\end{table}

\subsection{Models}

For all three tasks, we use large language models (LLMs) that can handle large documents efficiently. However, we also keep in mind two factors. First, the computational efficiency at inference time, as we aim for running the models on the entire dataset of 300,511 Czech courts' decisions in future work. Second, we try to avoid closed models to ensure replicability of our research \citep{Palmer2024}. We therefore experiment with two complementary state-of-the-art open architectures: the decoder-only Llama 3.1~8B~\citep{ai2024llama3} and the encoder-only ModernBERT~\citep{modernbert}. Llama 3.1 is trained on multilingual data, making it suitable for our Czech data, while ModernBERT provides a strong encoder baseline with long-context support. Moreover, for the holistic formalism classification task, we also experiment with feature-based multi-layer perceptron (MLP). For further details on the models, continued pretraining and hyperparameters, see Appendix~\ref{app.hyperparameters}.


\subsection{Results}

For all three tasks, we rely on a balanced evaluation based on the F1 score. To provide a more complete picture of model behavior, we distinguish between two complementary measures which we later average: 
the positive F1 score ($\text{F1}^+$), which measures how well the model identifies labels that are actually present, 
and the negative F1 score ($\text{F1}^-$), which measures how well the model avoids predicting labels that are absent. 
Standard F1 originated in information retrieval, where often only one class matters \citep{FMeasure2024}. In classification tasks like ours, where both the presence and absence of a label are often meaningful, reporting only positive-class F1 gives an incomplete picture as mentioned by \citet{FMeasure2024}. We therefore report both $\text{F1}^+$ and $\text{F1}^-$, and average them into a Balanced F1 score ($\text{Bal-F1}$) that penalizes models which perform well on one class at the expense of the other. We define $\text{Bal-F1}$ formally in the context of Task~2.

For \textbf{Task 2} (argument type classification), the problem is formulated as a multi-label multi-class classification task with an inventory of eight argument types:
\[
L = \{\text{LIN}, \text{SI}, \text{CL}, \text{D}, \text{HI}, \text{PL}, \text{TI}, \text{PC}\}.
\]
Following standard practice in multi-label evaluation, each label $\ell \in L$ is treated as an independent binary classification problem \citep{Tsoumakas2009, Sorower2010ALS}.

In our dataset, most argument types occur relatively sparsely, resulting in a strong imbalance between positive and negative instances. 
Standard multi-label metrics such as Hamming loss or positive-class F1 capture only part of the model behavior. 
In particular, positive-class F1 measures the ability of a model to detect argument types when they occur but ignores true negatives. 
However, for our purposes it is also important that models avoid predicting argument types that are not present. 
The goal of our paper is to develop models that will later be used for empirical legal analysis, where predictions are also used to estimate the frequency of argument types in large corpora. 
Systematic overprediction would artificially inflate these frequencies and therefore distort the resulting analysis.

Balanced evaluation metrics such as balanced accuracy explicitly incorporate both positive and negative predictions \citep{Borderson2010}. In analogy to this idea, we define a balanced variant of the F1 score that evaluates both classes symmetrically.

For each label $\ell$, we count:
\begin{itemize}
    \item \emph{true positives} ($\mathrm{tp}$): $\ell$ is annotated and predicted,
    \item \emph{false positives} ($\mathrm{fp}$): $\ell$ is predicted but not annotated,
    \item \emph{false negatives} ($\mathrm{fn}$): $\ell$ is annotated but not predicted,
    \item \emph{true negatives} ($\mathrm{tn}$): $\ell$ is neither annotated nor predicted.
\end{itemize}

From these counts we compute precision and recall for both the positive and negative class. 
The negative-class metrics are obtained by treating the negative class as the target class:

\begin{equation}
\text{Precision}^+_\ell = 
\frac{\mathrm{tp}_\ell}{\mathrm{tp}_\ell + \mathrm{fp}_\ell},
\qquad
\text{Recall}^+_\ell =
\frac{\mathrm{tp}_\ell}{\mathrm{tp}_\ell + \mathrm{fn}_\ell}
\end{equation}

\begin{equation}
\text{Precision}^-_\ell =
\frac{\mathrm{tn}_\ell}{\mathrm{tn}_\ell + \mathrm{fn}_\ell},
\qquad
\text{Recall}^-_\ell =
\frac{\mathrm{tn}_\ell}{\mathrm{tn}_\ell + \mathrm{fp}_\ell}
\end{equation}

Using these quantities, the positive and negative F1 scores are defined as:

\begin{equation}
\text{F1}^+_\ell =
\frac{2 \cdot \text{Precision}^+_\ell \cdot \text{Recall}^+_\ell}
{\text{Precision}^+_\ell + \text{Recall}^+_\ell},
\qquad
\text{F1}^-_\ell =
\frac{2 \cdot \text{Precision}^-_\ell \cdot \text{Recall}^-_\ell}
{\text{Precision}^-_\ell + \text{Recall}^-_\ell}.
\end{equation}

We then define the \textbf{Balanced F1} score for label $\ell$ as the macro-average of the positive and negative F1 scores:

\begin{equation}
\text{$\text{Bal-F1}$}_\ell =
\frac{\text{F1}^+_\ell + \text{F1}^-_\ell}{2}.
\end{equation}

Finally, the overall Macro $\text{Bal-F1}$ score is obtained by averaging the $\text{Bal-F1}$ score across all labels:

\begin{equation}
\text{Macro $\text{Bal-F1}$} =
\frac{1}{|L|} \sum_{\ell \in L} \text{$\text{Bal-F1}$}_\ell.
\end{equation}

To ensure transparency and comparability, we also report $\text{F1}^+$ and $\text{F1}^-$ for each argument type in Table~\ref{tab.task2.results}, and full per-class precision and recall ($\text{P}^+$, $\text{R}^+$, $\text{P}^-$, $\text{R}^-$) for each model in our repository.\footnote{\url{https://github.com/trusthlt/madon}} As we mentioned, the Balanced F1 metric complements $\text{F1}^+$ by additionally penalizing systematic overprediction, which is important when predictions are used for downstream empirical analysis of argumentation patterns.

For \textbf{Task 1} (argument presence detection) and \textbf{Task 3} (holistic formalism classification), 
the evaluation problem is binary: either an argument (resp.\ formalism) is present or not. 
We therefore compute F1$^+$ for the positive class (argument present/formalistic decision), F1$^-$ for the negative class (arguments absent/non-formalistic decision) and Bal-F1 that averages the two.  For these tasks, we again report per-class precision and recall ($\text{P}^+$, $\text{R}^+$, $\text{P}^-$, $\text{R}^-$) to provide a complete and transparent picture of model performance.

As illustrated in Table~\ref{tab:task3_main}, some models achieve relatively high $\text{F1}^+$ scores while having noticeably lower $\text{F1}^-$ for holistic classification of formalistic and non-formalistic decisions. On the other hand, Table~\ref {tab.task2.results} shows that for argument classification, the models achieve very high $\text{F1}^-$ but lower $\text{F1}^+$ scores. Since both the positive and negative classes are of interest, we believe that the Balanced F1 score provides important additional assessment of model performance.

\subsubsection{Results for Task 1: Argument presence detection}

\begin{table}
\centering
\caption{Results of argument presence detection (Task 1).
$\text{P}^+$/$\text{R}^+$/$\text{F1}^+$ are precision, recall, and F1 for the
positive class (argument present);
$\text{P}^-$/$\text{R}^-$/$\text{F1}^-$ for the negative class (argument absent).
$\text{Bal-F1}$ is the macro-averaged F1 over both classes.
Best $\text{Bal-F1}$ in bold.}
\label{tab:task1_main}
\scriptsize
\setlength{\tabcolsep}{3pt}
\begin{tabular}{@{}lrrrrrrrr@{}}
\toprule
\textbf{Model} & $\textbf{P}^+$ & $\textbf{R}^+$ & $\textbf{F1}^+$ & $\textbf{P}^-$ & $\textbf{R}^-$ & $\textbf{F1}^-$ & \textbf{$\text{Bal-F1}$} \\
\midrule
\multicolumn{8}{l}{\textit{Baselines}} \\
Majority baseline   & 0.0 & 0.0 & 0.0 & 87.1 & 100.0 & 93.1 & 46.6 \\
Random baseline     & 13.2 & 52.1 & 21.1 & 87.4 & 49.2 & 63.0 & 42.0 \\
\midrule
\multicolumn{8}{l}{\textit{Llama 3.1 8B}} \\
\llamabase          & 80.8 & 52.1 & 63.3 & 93.3 & 98.2 & 95.7 & 79.5 \\
\llamabase~+ PEFT   & 17.0 & 41.7 & 23.0 & 88.9 & 68.6 & 76.5 & 49.7 \\
\llamacpt           & 80.7 & 32.9 & 46.7 & 90.9 & 98.8 & 94.7 & 70.7 \\
\llamacpt~+ PEFT    & 18.8 & 43.3 & 25.9 & 88.6 & 68.0 & 76.7 & 51.3 \\
\midrule
\multicolumn{8}{l}{\textit{ModernBERT}} \\
\modernbertbase     & 77.8 & 62.1 & 69.1 & 94.6 & 97.4 & 96.0 & 82.5 \\
\modernbertcpt      & 77.4 & 62.9 & 69.3 & 94.7 & 97.3 & 96.0 & \cellcolor{shadecolor}\textbf{82.6} \\
\bottomrule
\end{tabular}
\end{table}

We report the results in Table~\ref{tab:task1_main}.

Models from ModernBERT family perform the best and reach $\text{Bal-F1}$ score around 82.5\%. Given that ModernBERT is a much smaller model compared to Llama 3.1 8B, its ability to slightly outperform the larger Llama model suggests that this task does not require the full capacity or extended context window of a large-scale LLM. This opens the possibility of deploying smaller, more efficient models in practice without sacrificing accuracy. Continued pretraining on Czech legal texts did not enhance performance for either architecture. In fact, for Llama, the CPT variant underperforms the base model (70.7\% vs.\ 79.5\% $\text{Bal-F1}$), and for ModernBERT the difference is negligible (82.6\% vs.\ 82.5\% $\text{Bal-F1}$). This suggests that the task can be solved effectively using base models.  

Surprisingly, parameter-efficient fine-tuning via LoRA led to a severe drop in performance. \llamabase~+ PEFT configuration fell from 79.5\% to 49.7\% $\text{Bal-F1}$, while the \llamacpt~+ PEFT variant performed similarly poorly at 51.3\%. This indicates that for argument presence detection, full fine-tuning is essential to achieve high performance.

\subsubsection{Results for Task 2: Argument type classification}
Table~\ref{tab.task2.results} presents results for each argument type and macro-averaged results for Task 2. The PEFT models only achieve macro $\text{F1}^+$ scores around random baseline (ranging from 2.9\% to 4.9\%). This indicates that parameter-efficient fine-tuning alone is insufficient for argument classification in our setting. In contrast, the best configuration of ModernBERT we used, \modernbertcpt~+ Asymmetric Loss, yields significantly better results, achieving a macro $\text{F1}^+$ of 44.4\%. This is far beyond the random baseline. However, the fully fine-tuned Llama~3.1~8B models achieve even better results than the ModernBERT family. The best-performing Llama~3.1~8B model is the base model trained with Asymmetric Loss, yielding a macro $\text{F1}^+$ score of 56.1\%, reaching a $\text{Bal-F1}$ of 77.5\%. 


This configuration consistently outperforms all baselines and other model setups, indicating that full fine-tuning allows the model to better capture the distribution and complexity of argument types in the MADON dataset. ModernBERT achieves competitive results, with its best performance coming from the CPT + Asy variant (balanced F1 of 71.6\%), though still notably behind the fully fine-tuned Llama. In contrast, PEFT-based Llama 3.1 8B configurations yield much lower $\text{Bal-F1}$ scores (49.6\% at best), performing only slightly above the majority baseline (49.4\%) and suggesting that the parameter-efficient setup, as implemented here, is again insufficient for this task.

Across all architectures, replacing Binary Cross-Entropy with Asymmetric Loss consistently boosts macro-average performance. The gain is most evident in ModernBERT (Base: 62.2\% $\rightarrow$ 67.3\%; CPT: 62.9\% $\rightarrow$ 71.6\%) and in the fully fine-tuned Llama 3.1 8B (74.6\% $\rightarrow$ 77.5\%). This pattern aligns with the class imbalance observed in the MADON dataset, where the negative class dominates across all labels, and Asymmetric Loss better emphasizes minority positive instances. 

The effect of continued pretraining (CPT) varies by architecture: for ModernBERT, \modernbertcpt~leads to substantial gains, while for Llama 3.1 8B full fine-tuning, \llamacpt~slightly underperforms the Base variant.

\paragraph{Error analysis}
Three factors most likely contribute to the limited performance on the hardest argument types. First, low inter-annotator agreement (IAA) likely introduces label noise that destabilizes training and imposes a performance ceiling. Practical Consequences and Systemic Interpretation have the lowest Krippendorff's alpha and rank among the worst-performing classes for both model families.

Second, label ambiguity complements this problem. Legal experts found that roughly half of all misclassifications for Historical Interpretation and Practical Consequences were reasonable from legal perspective. For Historical Interpretation, frequent borderline cases involved EU legislative preambles that include the aim of the legislation and counting reference to them as Historical Interpretation is contentious. For Practical Consequences, models often predicted this label when courts discussed consequences as a precondition for applying a legal rule rather than as an interpretive argument. While such cases were excluded from the gold standard, the model's predictions are understandable from the legal perspective given the genuine structural similarity between the two functions practical consequences can serve in legal reasoning.

Third, label frequency matters. Historical Interpretation is both the least frequent class (58 paragraphs) and among the hardest for both model families. However, this issue does not apply universally. Doctrine has similarly few examples (68 paragraphs) but performs well---likely because its high IAA compensates for sparsity.

\begin{table}
\centering
\scriptsize
\setlength{\tabcolsep}{2.8pt}
\caption{\label{tab.task2.results} Task 2: Metrics per each argument type showing $\text{F1}^+$, $\text{F1}^-$ and $\text{Bal-F1}$ for each argument and macro average for all eight arguments. BCE = Binary Cross-Entropy, Asy = Asymmetric Loss, PEFT = Parameter-Efficient Fine-Tuning, CPT = Continued Pretrained Model. Bold means best Macro $\text{Bal-F1}$.}
\begin{tabular}{p{0.6cm} p{3.0cm} l *{8}{r} r}
\toprule
& \textbf{Model} & \textbf{Metrics} & \textbf{LIN} & \textbf{SI} & \textbf{CL} & \textbf{D} & \textbf{HI} & \textbf{PL} & \textbf{TI} & \textbf{PC} & \textbf{Macro} \\
\midrule

\multirow{6}{*}{\rotatebox[origin=c]{90}{\textbf{Baselines}}}
& \multirow{3}{3.1cm}{\textbf{Majority}} &
F1$^+$ & 0.0 & 0.0 & 0.0 & 0.0 & 0.0 & 0.0 & 0.0 & 0.0 & 0.0 \\
& & F1$^-$ & 99.4 & 98.8 & 96.6 & 99.9 & 99.6 & 97.7 & 98.3 & 99.2 & 98.7 \\
& & Bal-F1 & 49.7 & 49.4 & 48.3 & 49.9 & 49.8 & 48.9 & 49.2 & 49.6 & 49.4 \\
\cmidrule(lr){2-12}
& \multirow{3}{3.1cm}{\textbf{Random}} &
F1$^+$ & 2.2 & 4.4 & 12.7 & 0.8 & 1.5 & 7.4 & 6.1 & 1.8 & 4.6 \\
& & F1$^-$ & 66.5 & 64.2 & 66.9 & 68.0 & 67.2 & 67.1 & 65.2 & 65.8 & 66.3 \\
& & Bal-F1 & 34.3 & 34.3 & 39.8 & 34.4 & 34.3 & 37.2 & 35.7 & 33.8 & 35.5 \\
\midrule

\multirow{12}{*}{\rotatebox[origin=c]{90}{\textbf{Llama 3.1 8B (PEFT)}}}
& \multirow{3}{3.1cm}{\textbf{Base + BCE}} &
F1$^+$ & 0.8 & 4.3 & 17.6 & 0.0 & 3.4 & 6.1 & 4.4 & 2.7 & 4.9 \\
& & F1$^-$ & 94.2 & 92.1 & 80.6 & 92.3 & 93.3 & 79.2 & 84.7 & 88.1 & 88.1 \\
& & Bal-F1 & 47.5 & 48.2 & 49.1 & 46.1 & 48.4 & 42.7 & 44.5 & 45.4 & 46.5 \\
\cmidrule(lr){2-12}
& \multirow{3}{3.1cm}{\textbf{Base + Asy}} &
F1$^+$ & 0.0 & 1.6 & 18.9 & 0.0 & 0.9 & 3.3 & 1.5 & 3.1 & 3.7 \\
& & F1$^-$ & 97.5 & 97.4 & 90.8 & 96.4 & 97.3 & 94.0 & 94.6 & 96.4 & 95.6 \\
& & Bal-F1 & 48.8 & 49.5 & 54.9 & 48.2 & 49.1 & 48.6 & 48.1 & 49.8 & 49.6 \\
\cmidrule(lr){2-12}
& \multirow{3}{3.1cm}{\textbf{CPT + BCE}} &
F1$^+$ & 0.0 & 6.2 & 17.4 & 0.0 & 2.3 & 7.2 & 2.1 & 0.7 & 4.5 \\
& & F1$^-$ & 97.1 & 93.6 & 82.2 & 91.1 & 92.2 & 83.5 & 90.0 & 95.7 & 90.7 \\
& & Bal-F1 & 48.5 & 49.9 & 49.8 & 45.6 & 47.2 & 45.4 & 46.0 & 48.2 & 47.6 \\
\cmidrule(lr){2-12}
& \multirow{3}{3.1cm}{\textbf{CPT + Asy}} &
F1$^+$ & 0.0 & 1.1 & 13.7 & 0.0 & 2.4 & 2.3 & 1.6 & 1.7 & 2.9 \\
& & F1$^-$ & 98.7 & 97.5 & 90.9 & 97.8 & 98.2 & 92.9 & 96.2 & 98.4 & 96.3 \\
& & Bal-F1 & 49.3 & 49.3 & 52.3 & 48.9 & 50.3 & 47.6 & 48.9 & 50.0 & 49.6 \\
\specialrule{0.6pt}{0pt}{0pt}

\multirow{12}{*}{\rotatebox[origin=c]{90}{\textbf{Llama 3.1 8B (Full)}}}
& \multirow{3}{3.1cm}{\textbf{Base + BCE}} &
F1$^+$ & 41.4 & 30.3 & 81.7 & 75.0 & 14.1 & 75.4 & 48.2 & 34.7 & 50.1 \\
& & F1$^-$ & 99.5 & 98.9 & 98.8 & 99.9 & 99.6 & 98.9 & 98.6 & 99.4 & 99.2 \\
& & Bal-F1 & 70.4 & 64.6 & 90.2 & 87.5 & 56.9 & 87.1 & 73.4 & 67.0 & 74.6 \\
\cmidrule(lr){2-12}
& \multirow{3}{3.1cm}{\textbf{Base + Asy}} &
F1$^+$ & 57.6 & 34.8 & 79.5 & 63.9 & 36.2 & 74.8 & 55.1 & 46.9 & 56.1 \\
& & F1$^-$ & 99.4 & 98.3 & 98.5 & 99.8 & 99.6 & 98.7 & 98.2 & 99.1 & 99.0 \\
& & Bal-F1 & 78.5 & 66.5 & 89.0 & 81.9 & 67.9 & 86.8 & 76.7 & 73.0 & \cellcolor{shadecolor}\textbf{77.5} \\
\cmidrule(lr){2-12}
& \multirow{3}{3.1cm}{\textbf{CPT + BCE}} &
F1$^+$ & 36.8 & 16.6 & 81.0 & 75.0 & 0.0 & 75.8 & 51.4 & 26.2 & 45.3 \\
& & F1$^-$ & 99.5 & 98.8 & 98.7 & 99.9 & 99.6 & 98.9 & 98.7 & 99.3 & 99.2 \\
& & Bal-F1 & 68.1 & 57.7 & 89.9 & 87.5 & 49.8 & 87.4 & 75.0 & 62.8 & 72.3 \\
\cmidrule(lr){2-12}
& \multirow{3}{3.1cm}{\textbf{CPT + Asy}} &
F1$^+$ & 56.3 & 37.0 & 78.2 & 67.2 & 19.4 & 70.7 & 57.6 & 43.0 & 53.7 \\
& & F1$^-$ & 99.4 & 98.4 & 98.3 & 99.9 & 99.5 & 98.5 & 98.3 & 99.0 & 98.9 \\
& & Bal-F1 & 77.8 & 67.7 & 88.2 & 83.5 & 59.4 & 84.6 & 77.9 & 71.0 & 76.3 \\
\specialrule{0.6pt}{0pt}{0pt}

\multirow{12}{*}{\rotatebox[origin=c]{90}{\textbf{ModernBERT}}}
& \multirow{3}{3.1cm}{\textbf{Base + BCE}} &
F1$^+$ & 9.6 & 5.9 & 69.7 & 24.4 & 9.3 & 49.1 & 27.7 & 10.0 & 25.7 \\
& & F1$^-$ & 99.2 & 98.5 & 98.0 & 99.8 & 99.4 & 98.1 & 98.0 & 98.9 & 98.8 \\
& & Bal-F1 & 54.4 & 52.2 & 83.9 & 62.1 & 54.3 & 73.6 & 62.9 & 54.4 & 62.2 \\
\cmidrule(lr){2-12}
& \multirow{3}{3.1cm}{\textbf{Base + Asy}} &
F1$^+$ & 19.4 & 17.8 & 67.5 & 67.2 & 23.3 & 48.5 & 32.2 & 11.4 & 35.9 \\
& & F1$^-$ & 99.3 & 98.3 & 97.5 & 99.9 & 99.6 & 97.8 & 97.7 & 99.0 & 98.6 \\
& & Bal-F1 & 59.3 & 58.0 & 82.5 & 83.5 & 61.5 & 73.1 & 65.0 & 55.2 & 67.3 \\
\cmidrule(lr){2-12}
& \multirow{3}{3.1cm}{\textbf{CPT + BCE}} &
F1$^+$ & 9.3 & 13.3 & 67.4 & 28.6 & 3.4 & 49.2 & 37.4 & 12.7 & 27.7 \\
& & F1$^-$ & 97.0 & 97.8 & 97.8 & 98.8 & 98.7 & 98.2 & 97.7 & 98.3 & 98.0 \\
& & Bal-F1 & 53.2 & 55.5 & 82.6 & 63.7 & 51.1 & 73.7 & 67.5 & 55.5 & 62.9 \\
\cmidrule(lr){2-12}
& \multirow{3}{3.1cm}{\textbf{CPT + Asy}} &
F1$^+$ & 29.9 & 32.7 & 75.1 & 59.3 & 21.9 & 63.2 & 39.5 & 33.6 & 44.4 \\
& & F1$^-$ & 99.1 & 98.4 & 98.2 & 99.8 & 99.3 & 98.1 & 98.0 & 99.2 & 98.8 \\
& & Bal-F1 & 64.5 & 65.6 & 86.7 & 79.5 & 60.6 & 80.6 & 68.7 & 66.4 & 71.6 \\
\bottomrule
\end{tabular}
\end{table}

\subsubsection{Results for Task 3: Holistic formalism classification} 
\begin{table}
\centering
\caption{Results of holistic formalism classification (Task~3).
$\text{P}^+$/$\text{R}^+$/$\text{F1}^+$ are precision, recall, and F1 for the
positive class (formalistic);
$\text{P}^-$/$\text{R}^-$/$\text{F1}^-$ for the negative class (non-formalistic).
$\text{Bal-F1}$ is the macro-averaged F1 over both classes.
Bold = highest $\text{Bal-F1}$ per block.}
\label{tab:task3_main}
\scriptsize
\setlength{\tabcolsep}{3pt}
\begin{tabular}{@{}lrrrrrrrr@{}}
\toprule
\textbf{Model} & $\textbf{P}^+$ & $\textbf{R}^+$ & $\textbf{F1}^+$ & $\textbf{P}^-$ & $\textbf{R}^-$ & $\textbf{F1}^-$ & \textbf{$\text{Bal-F1}$} \\
\midrule
\multicolumn{8}{l}{\textit{Baselines}} \\
Majority baseline               & 58.6 & 100.0 & 73.9 & 0.0 & 0.0 & 0.0 & 37.0 \\
Random baseline                 & 55.6 & 29.4 & 38.5 & 40.0 & 66.7 & 50.0 & 44.2 \\
\midrule
\multicolumn{8}{l}{\textit{Gold-annotated arguments as features}} \\
MLP                             & 92.4 & 94.1 & 93.2 & 91.4 & 88.9 & 90.1 & \cellcolor{shadecolor}\textbf{91.7} \\
\midrule
\multicolumn{8}{l}{\textit{End-to-end models}} \\
\llamabase                      & 78.1 & 84.3 & 81.0 & 76.1 & 66.7 & 70.8 & 75.9 \\
\llamabase~+ PEFT               & 64.5 & 96.1 & 77.2 & 80.6 & 25.0 & 38.1 & 57.6 \\
\llamacpt                       & 79.1 & 88.2 & 83.3 & 80.7 & 66.7 & 72.6 & 78.0 \\
\llamacpt~+ PEFT                & 65.9 & 98.0 & 78.8 & 93.3 & 27.8 & 41.9 & 60.3 \\
\modernbertbase                 & 78.7 & 78.4 & 78.5 & 69.5 & 69.4 & 69.2 & 73.8 \\
\modernbertcpt                  & 84.6 & 86.3 & 85.4 & 80.1 & 77.8 & 78.9 & \cellcolor{shadecolor}\textbf{82.2} \\
\midrule
\multicolumn{8}{l}{\textit{Multi-step pipelines with MLP}} \\
No Filtering                    & 85.3 & 88.2 & 86.7 & 82.1 & 77.8 & 79.7 & 83.2 \\
With Filtering                  & 81.8 & 96.1 & 88.3 & 92.6 & 69.4 & 79.2 & \cellcolor{shadecolor}\textbf{83.8} \\
\bottomrule
\end{tabular}
\end{table}

\paragraph{Gold labeled arguments as features}

Arguments are crucial for holistic evaluation of formalism. MLP reaches $\text{Bal-F1}$ of 91.7\% in classifying decisions when we used human expert annotations of arguments from MADON dataset as “oracle" features. This performance is substantially higher than any of our end-to-end LLMs (see Table~\ref{tab:task3_main}). The comparison is obviously not fair, as the input of LLMs is the entire decision plain text while the MLP knows the true arguments, but it shows that the presence, absence, and distribution of specific argument types do capture the essence of judicial formalism, even though the legal experts considered other aspects (e.g., context of the case) when assigning the holistic label for the MADON dataset. This upper bound is optimistic, as it assumes perfect argument labels.

\begin{figure}
\subcaptionbox{Non-formalistic label}[.5\linewidth]%
{\includegraphics[width=\linewidth,trim={0.65cm 0.85cm 1.6cm 0.8cm},clip]{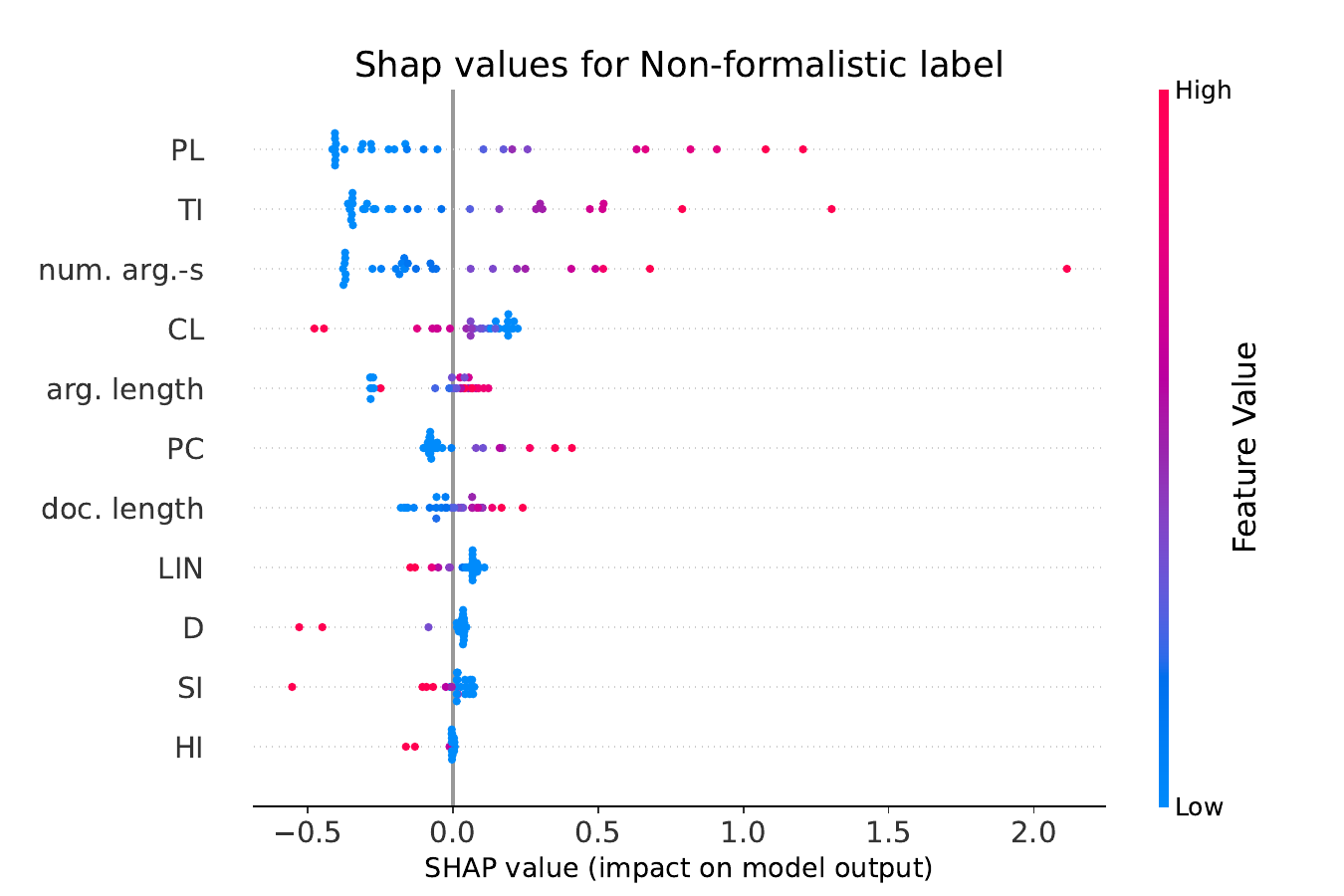}}
\hspace*{\fill}
\subcaptionbox{Formalistic label}[.5\linewidth]%
{\includegraphics[width=\linewidth,trim={0.65cm 0.85cm 1.5cm 0.8cm},clip]{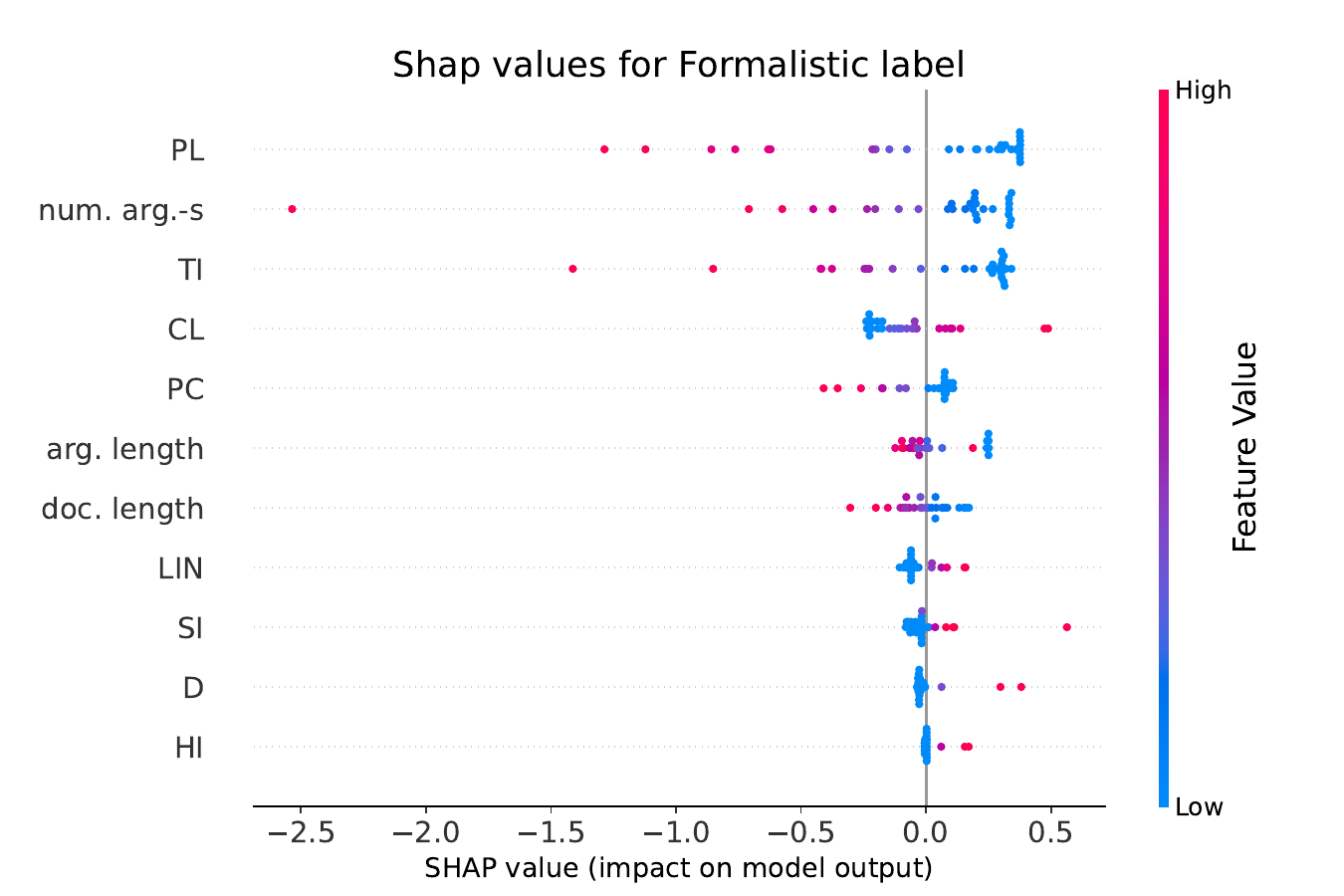}}
\caption{Summary plot for SHAP values of all test examples}
\label{fig:shap_summary1}
\end{figure}

Not all arguments contribute equally. Figure~\ref{fig:shap_summary1} summarizes an explainability method using SHAP analysis \citep{lundberg2017unified} and shows that Principles of Law and Values (PL) arguments are the strongest predictor of non-formalism: higher frequencies of PL are closely linked to non-formalistic decisions. Teleological Interpretation (TI) and the total number of arguments also correlate positively with non-formalism, which aligns with our theoretical definition that describes formalism as the avoidance of purposive reasoning and limited number of arguments. By contrast, Systemic Interpretation (SI), though classified as formalistic in theory, has little predictive value.

\paragraph{Contemporary end-to-end LLMs for holistic classification}

Direct classification of decisions as formalistic/non-formalistic (i.e., without argument features) gives very promising results. Continued pretraining improves both model families, and  ModernBERT-large-Czech-Legal achieves 82.2\% $\text{Bal-F1}$, the best among end-to-end models. ModernBERT outperforms Llama even though Llama can process the entire document context without truncating the text after 8,192 tokens, unlike ModernBERT. This indicates that lexical and semantic cues, which also play an important role in expert annotation, may be more valuable than full-context processing. ModernBERT outperforms the baselines by 40+ points, yet still falls behind the argument-based approach by about ten percentage points.

\paragraph{Multi-step pipeline for holistic classification}
Since even the best LLMs fall behind the feature-based performance, we also designed a multi-step pipeline that uses features and mirrors how legal scholars would evaluate a decision and assess formalism: first filtering irrelevant passages, then identifying argument types, and finally assigning a holistic label formalistic/non-formalistic. This architecture provides better explainability and even slightly improves performance. The pipeline consists of three stages:
\begin{enumerate}
\item In the first stage, ModernBERT Large identifies which paragraphs contain legal arguments, filtering out ca 85\% of the decisions' text (i.e., non-argumentative parts of the decisions; these are often repetitive or procedural texts usually not crucial for assessing formalism). 
\item In the second stage, Llama 3.1 8B Base (+ Asy), classifies the filtered text into argument types from the argument inventory (recall our taxonomy of eight legal arguments in Section~\ref{subsec:measure_formalism}).
\item Finally, in the last step with MLP, we extract features from the document containing the automatically annotated arguments from Llama focusing mainly on the argument distribution. MLP then classifies the input as formalistic or non-formalistic. We rely on our previous finding that argument frequency strongly predicts the overall formalistic/non-formalistic nature of the decision.
\end{enumerate}

The pipeline with filtering (83.8\% Bal-F1) slightly outperforms both the best end-to-end model (82.2\%) and the simpler pipeline without filtering (83.2\%) as depicted by Table~\ref{tab:task3_main}.  These results provide two key findings. First, decomposing NLP classification task according to legal theory does not sacrifice performance; on the contrary, the pipeline scores slightly better than end-to-end models and provides greater explainability (due to the reliance on features in MLP that can be explained by SHAP methods). Second, this approach can save compute costs (as filtering as much as 85\% of the non-argumentative text reduces the amount of text to be classified with Llama inference).

\section{Discussion}

Our study provides three main findings. First,  even complex, contested legal concept such as formalism can be detected in non-English court decisions using NLP (Bal-F1 of 83.8\%). Second, we show that decomposing legal classification tasks and using more explainable, segmented pipelines across model families can improve performance and explainability while reducing compute costs. Third, the first empirical results weaken dominant narratives about judicial formalism in Central and Eastern Europe, particularly in Czechia.

\subsection{Contributions to legal NLP and computational legal studies}
This article provides the first LLM-based argument mining study on CEE courts. Our experiments and lessons learned on non-English court judgments could benefit the legal NLP community and legal researchers in other CEE states. Parameter-efficient fine-tuning was inadequate, whereas continued pretraining proved important for ModernBERT (namely for argument-type classification and holistic formalism classification). By contrast, Llama benefited primarily from full fine-tuning rather than continued pretraining. Class imbalance, which often prevails in legal datasets, needs to be reflected as suggested by \citet{wais2026}. For multi-label argument-type classification, Asymmetric Loss consistently outperformed Binary Cross-Entropy across model architectures, improving Bal-F1 by up to nine points. Finally, macro-averaged metrics were essential for multi-label evaluation; the scores would be otherwise dominated by the majority negative class for each argument type.

Additionally, we show that the interdisciplinary combination of legal theory and NLP can be fruitful. Rather than classifying reasoning style at the decision level directly, this NLP task benefits from decomposing the pipeline according to legal theory, i.e., first, filter legally relevant parts of the text, then classify traditional legal arguments, and only then classify legal formalism. We expect that decomposing classification tasks in accordance with how lawyers or legal theorists approach problems could boost explainability and performance in other legal NLP tasks.

This article also contributes to the “law as data" strand of computational legal studies \citep{Chau_Livermore_2024}, which uses computational tools to study courts and judicial culture at scale. We show that legal argument mining, even when imperfect, enables promising classification of judicial philosophies; noisy classification of some arguments can still capture the key dimensions of judicial reasoning with sufficient performance for downstream empirical analysis. As we have shown, this can then challenge entrenched narratives and provide systematic evidence where anecdote previously dominated. We release the MADON dataset, the accompanying corpus of 300,511 decisions, as well as all the models to provide the necessary resources for this research agenda.

\subsection{Implications for legal scholarship}

We provide a case study from Czechia that challenges some aspects of the CEE formalism narrative. Both the Supreme Court and the Supreme Administrative Court use textual arguments scarcely (5.7\% of all arguments) and include teleological or value-based reasoning much more frequently than textualist reasoning. Since this finding also applies to the SC, i.e., to the less reformed court with institutional ties to the previous regime, it challenges the claim that unreformed courts avoid non-formalistic reasoning and are text-bound \citep{kuhn_judiciary_2011-2}. We thus confirm the observation of \citet{cserne_judicial_2024} about the lack of evidence: for textualism, the key tenet of the alleged formalism, empirical support is still missing. 

The very low frequency of text-based reasoning is also surprising in comparative perspective, especially in the context of the recent US research. While any comparisons must be taken with a pinch of salt, there is a strong consensus supported by substantial empirical evidence that textualism had been rising since the 1980s and is now the important interpretive approach at the US Supreme Court and US District Courts \citep{choi_2020}, as well as US State Supreme Courts \citep{Peters2024}. Our findings suggest the opposite is true for Czech courts. Contrary to expectations, they scarcely engage with the text of statutes when applying them. Our findings align with similar research on German criminal cases, where the frequency of linguistic interpretation was also rather low---more than half of the cases did not contain any reference to statutory wording, even though the authors expected that criminal law, governed by the strict legality principle, would cite wording more frequently \citep{KudlichChristensen2009}.

Additionally, case law dominates the reasoning of both Czech supreme courts and appears in more than half of the oldest decisions in our dataset (1997-1998), which weakens the findings that some judiciaries within the CEE legal cultures like the Czech one had been to great extent precedent-averse in the 2000s and before \citep{Kuhn2015Precedent, Hondius2007}. 

Our findings speak directly only to the two Czech apex courts, and 
generalization requires caution.  It remains an open empirical question whether similar patterns hold across  other Czech courts or CEE jurisdictions. That said, existing scholarship provides reasons to treat the Czech case as informative for the broader region. Comparative studies classify all CEE countries as formalistic. While they note somewhat greater non-formalistic reasoning by administrative courts in Czechia than in Poland or Hungary \citep{matczak_constitutions_2010-2}, they still portray the Czech SC specifically as a typical post-communist apex court \citep{matczak_eu_2015}. Košař and Šipulová likewise use Czechia as a case for exploring judicial reforms across CEE \citep{sipulova_kosar_purging_2024-1}. In short, the literature often treats Czechia, and especially the Czech SC, as representative of the wider CEE pattern. The narrative about CEE is thus doubtful as it lacks broader evidence and seems to fail for one of its own paradigm cases.

\subsection{Limitations and future directions}

Our research has several limitations.
First, empirical findings are restricted to 272 which, while representative thanks to various sampling methods, limits generalizability. Nonetheless, this issue will be addressed by the large scale analysis with our models we plan to pursue as a next step. 

Second, lower inter-annotator agreement for some argument types limits reproducibility and partly constrains F1 scores on those categories. Notably, many annotator disagreements and many model misclassifications are reasonable, in the sense that good arguments existed for the label that was ultimately rejected. This connects to a broader question in NLP about whether a single gold-standard label is desirable for supervised learning at all, with growing skepticism across many NLP subfields \citep{Perspectivist40, plank-2022-problem} and with some scepticism in legal domain \citep{braun_i_2023-1}.  We have been encountering reasonable disagreements repeatedly in our legal argument mining work \citep{Habernal.et.al.2023.AILaw,Held.Habernal.2024.AILaw} and increasingly see label variation as “an opportunity rather than a problem" \citep{plank-2022-problem}. Possible responses include retaining reasonable disagreements rather than forcing consensus, training on unaggregated annotations, training ensemble models, or adopting soft evaluation metrics \citep{rizzi-etal-2024-soft}. In this paper, we take a first step by going beyond the standard in legal NLP and publishing the raw annotations including disagreements \citep{braun_i_2023-1}.

Third, model evaluation relies on a small test set (29 decisions). Some models will later be applied to much larger datasets (e.g., 300,000 decisions of 10 Czech courts in our corpus) and used for downstream statistical analysis (e.g., for estimating argument frequencies in all decisions ever issued). Reliable deployment therefore requires additional validation, confidence intervals as well as possible error correction \citep{maranca2025correcting} or sensitivity analysis.

Lastly, the sparse occurrence of text-based arguments (5.7\%) partially reflects our research design choices in our annotation guidelines. We required courts to engage at least partially with statutory language--analyzing wording, referencing dictionaries, or discussing grammar--rather than merely citing provisions. However, our threshold remained low. Any reference to wording or explicit textual analysis qualified as linguistic interpretation. If we counted every single reference to a statute as linguistic interpretation, the frequency would be much higher; however, we believe our approach aligns better with legal argumentation theory, which treats analyzing wording, referencing dictionaries, and discussing grammar as the core aspects of linguistic interpretation.

Our research could be expanded in two main ways. We are currently applying the trained models to the full corpus of 300,000 apex-court decisions to validate our empirical results and to test whether the models transfer to lower court decisions. Secondly, we believe our methodology could be extended to other jurisdictions, both in CEE and beyond, given that our argument taxonomy is grounded in general theories of legal argumentation \citep{maccormick_interpreting_2016, Walton_Macagno_Sartor_2021, alexy_argument_2010} and that models might transfer across languages and domains \citep{savelka2021, yeginbergen-etal-2024-argument, toledo-ronen-etal-2020-multilingual}. This would directly test whether Czech patterns are country-specific and probe contested theoretical claims about formalism.

\section{Conclusion}

We asked whether state-of-the-art NLP methods can detect argumentative paragraphs and classify them into eight traditional legal argument types in Czech judicial decisions, and whether they can distinguish between formalistic and non-formalistic reasoning styles at the decision level. Our results show they mostly can. After designing annotation guidelines for formalism and arguments and annotating 272 Czech Supreme Court decisions (9,183 paragraphs) resulting in the new MADON dataset, our extensive set of experiments shows that ModernBERT can distinguish argumentative from non-argumentative paragraphs with a $\text{Bal-F1}$ of 82.6\%. Argument-type classification achieves a promising $\text{Bal-F1}$ of 77.5\% using fully fine-tuned Llama 3.1 with asymmetric loss, though three argument types with low inter-coder agreement or low prevalence (historical interpretation, systemic interpretation, and practical consequences) remain difficult for models to classify. Holistic formalism classification reaches a very promising $\text{Bal-F1}$ of 83.8\% thanks to a novel three-stage pipeline that increases explainability and saves compute costs by first filtering the non-argumentative paragraphs (more than 85\% of all paragraphs) using a smaller model; given the high difficulty of the task for legal experts, we consider the F1 score very good. Empirically, we provide first evidence that text-based arguments are surprisingly rare in both courts, while case law and purposive and value-based reasoning dominate. The two courts' argumentation profiles seem to diverge after 2011, with the SAC trending more significantly toward non-formalistic reasoning. 

\section*{Acknowledgements}
This work has been supported by the German Research Foundation as part of the ECALP project (HA 8018/2-1), by the Research Center Trustworthy Data Science and Security (\url{https://rc-trust.ai}), one of the Research Alliance centers within the \url{https://uaruhr.de}, by the Charles University Grant Agency (project no.\ 185023), and by the Sylff Scholarship. We would like to thank Marlene Anzenberger, Kevin Ashley, Michal Bobek, Christoph Burchard, Zdenek Cervinek, Jonathan Choi, Corina Coupette, Dirk Hartung, Matthias Klatt, Pavel Ondrejek, Michael Preisig, Hubert Rottleuthner, Jaromir Savelka, Holger Spamann, Jed Stiglitz, and Barbara Zeller for their feedback on the early versions of this project, as well as all others who provided comments along the way. We also thank Czech Supreme Court and Supreme Administrative Court for their kind collaboration and answers with regards to their case law and databases. We are grateful to two independent reviewers of this journal for their excellent feedback and community service that led to major improvements of the initial draft. Special thanks to Václav Lipš, Vítek Eichler, Matyáš Barták, Marek Švajda for their tremendous research assistance. 

\bibliography{bibliography}

\begin{appendices}

\section{Arguments}\label{secA1}

For each argument type, our guidelines included a structured description that connects the theoretical concept to real‑world decisions. For most types, we (1) give a brief description, (2) list its subtypes, (3) note “trigger” phrases that usually signal its presence in a decision, (4) provide typical and borderline examples, (5) add comments and annotator tips and (6) include annotation chart.

All details are in the guidelines. What follows is a brief description intended to illustrate how we understand the particular argument types (\emph{all examples shown in italics}).

\subsection*{Linguistic Interpretation (LIN)}

This argument type focuses on the literal text of a legal provision. This includes analyzing the ordinary meaning of words, explicitly reflecting syntax or grammar rules, referencing explicit legal definitions within a statute, and employing reasoning a contrario. We considered trigger words like ``unambiguously worded,'' ``diction,'' or ``formulation'' as strong indicators.
\begin{quote}
\emph{``In addition to the above, the Supreme Administrative Court adds that the wording of Section 87e(1)(i)(1) of the Act on the Residence of Aliens is very unambiguous and leaves no room for a different interpretation.''}
\end{quote}

\subsection*{Systemic Interpretation (SI)}

This argument involves interpreting a provision by placing it within the broader context of the legal system to ensure a “logical, consistent and non-contradictory interpretation". This includes applying the collision rules (\textit{lex specialis, lex posterior, lex superior}), narrowly construing exceptions, and explicitly performing conforming interpretation (either constitutional conforming or EU conforming interpretation) to align domestic law with higher-order legal norms.
\begin{quote}
\emph{``Since the provisions of Section 281 of the Criminal Procedure Code do not contain special provisions for decisions pursuant to Section 288(3) of the Criminal Procedure Code, the general provisions on the subject matter and local jurisdiction of the court (Sections 16 to 22 of the Criminal Procedure Code) apply.''}
\end{quote}

\subsection*{Case Law (CL)}

This category captures express reliance on specific prior judicial decisions from domestic courts, the CJEU, or the ECHR. For an annotation to be made, the court must cite a particular case; a vague reference to “according to the case law" without a specific citation was not considered sufficient.
\begin{quote}
\emph{``In support of the conclusion, the Court also refers to the judgment of the Supreme Court of 11 May 2005, Case No. 30 Cdo 64/2004 (available at www.nsoud.cz).''}
\end{quote}

\subsection*{Doctrine (D)}

This argument type refers to citations of academic or expert commentary, such as legal textbooks, scholarly articles, or commentaries. References to legal dictionaries are also marked as Doctrine, whereas non-legal dictionaries fall under Linguistic Interpretation.
\begin{quote}
\emph{``Current commentary literature on both the Code of Civil Procedure and the Civil Code derives from the nature of a power of attorney that the attorney‑in‑fact’s signature is not among its formal requirements (Smolík, P., § 28 [Granting a Power of Attorney and its Termination], in: Svoboda, K. et al., Občanský soudní řád, 3rd ed., Prague: C. H. Beck, 2021, pp. 121–122).''}
\end{quote}

\subsection*{Historical Interpretation (HI)}

This argument type seeks to determine the lawmaker’s intent behind a provision. When court seeks to determine the intent, it typically refers to historical documents for legislative process, such as explanatory memoranda, stenographic records of parliamentary debates, or the circumstances of a law’s adoption. Due to its similarity, we also included arguments based on the assumption of a rational lawmaker.
\begin{quote}
\emph{``cf. the wording of (...) the special part of the explanatory report [``On points 4 and 5 (section 274)''] to the governmental bill No. 31/2019 Coll., which was debated by the Chamber of Deputies of the Parliament of the Czech Republic in its 8th term (2017-2021) as Document No. 71.''}
\end{quote}

\subsection*{Principles of Law and Values (PL)}

This non-formalistic category includes appeals to higher-order norms that transcend the literal text of a single rule. It covers general legal principles (e.g., good faith, legitimate expectations, legal certainty) and explicit references to fundamental rights and values found in the Czech Constitution (incl. the European Convention on Human Rights) or EU Law. A principle does not need to be explicitly labeled as such by the court to be annotated.
\begin{quote}
\emph{``[...] which constitutes an interference with the right to a fair trial enshrined in Article 36(1) of the Charter''}
\end{quote}

\subsection*{Teleological Interpretation (TI)}

This argument focuses on the purpose, goal, or “telos" of a legal provision, often signaled by terms like “purpose," “aim," “sense," or “function". This category also includes special forms of purpose-driven reasoning, such as arguments ad absurdum (to avoid a nonsensical result), a fortiori, and reasoning by analogy.
\begin{quote}
\emph{``An appeal is conceived as an extraordinary remedy and is therefore intended to remedy only serious legal defects in final decisions.''}
\end{quote}

\subsection*{Practical Consequences (PC)}

This argument type involves the court considering the real-world effects and practical outcomes of its decision. Instead of focusing on the wording of legal norms, the court weighs what will happen if it chooses one interpretation over another, answering the question “What will it lead to?". This can include impacts on the parties as well as the society.
\begin{quote}
\emph{In the event of a contrary conclusion, it would be difficult to clearly identify and distinguish what personal manifestations of the taxi driver in the course of his interaction with the customer can still be characterized as his work activity and what already constitutes his private life.''}
\end{quote}

\section{Models, continued-pretraining and hyperparameters}
\label{app.hyperparameters}
\subsection{Llama}
We use two variants of Llama 3.1~8B: the base model without instruction fine-tuning, which is suited for multilabel task finetuning, and a base model version which received continued pretraining (CPT) on Czech legal texts (Section~\ref{sec.continued.pretraining} below).
We also test the performance of instruction tuning (IT) on Llama 3.1~70B with different variants of prompting using annotation guidelines in zero-shot and few-shot settings. To optimize computational costs, we also test the performance of parameter-efficient fine-tuning (PEFT, via LoRA adapters~\citep{hu2021lora}). However, we find that instruction tuning performed significantly worse than fully fine-tuned models.

\subsection{ModernBERT}

ModernBERT~395M~\citep{modernbert} is a recent encoder-only transformer model supporting up to 8,192 tokens. Compared to earlier models of the BERT family, it integrates modifications such as Rotary Positional Embeddings (RoPE)~\citep{su-etal-2023-roformer}, GeGLU activations, alternating local–global attention, and Flash Attention~\citep{dao-etal-2022-flashattention}. These features make it a suitable encoder baseline for our classification tasks, overcoming the token limitation and English bias of Legal-BERT \citep{chalkidis2020legalbert}.

\subsection{Continued pretraining of transformers}
\label{sec.continued.pretraining}

Neither Llama 3.1~8B nor ModernBERT was originally trained for the legal domain or the Czech language specifically. To address this limitation, we perform continued pretraining of both models using Czech legal data. This approach allows us to investigate whether domain- and language-specific adaptation improves performance, building on previous findings in legal NLP~\citep{zheng2021whendoes, Habernal.et.al.2023.AILaw}.
For pretraining, we curate the corpus of approximately 300,000 legal decisions, excluding all decisions already included in the MADON dataset to prevent data contamination.\footnote{Data available at \url{https://github.com/trusthlt/madon}.}

Due to computational constraints, we continue pretraining the Llama 3.1 8B base model for one epoch, using an effective batch size of 16 and a maximum input length of 32,000 tokens, which is large enough to fully cover most of the documents. We use unsloth\footnote{https://unsloth.ai} to optimize the training process.

We follow a two-phase pretraining process for ModernBERT. First, we train a custom tokenizer using byte-pair encoding on our pretraining dataset. We then continue pretraining from the last checkpoint of ModernBERT-large using a masked language modeling objective.

In phase 1, we include documents with fewer than 8,192 tokens (the model’s input limit), train with a 30\% masking probability and an effective batch size of 32. In phase 2, we reduce the effective batch size to 8 and the masking probability to 15\% and train using the remaining documents.

\subsection{Multi-layer perceptron}

For holistic formalism classification, we also experiment with an alternative input representation. We simply use the annotated argument types and extract various statistics into a document-level feature vector which we feed into a multi-layer perceptron (MLP; 2 hidden layers with 20 and 50 neurons, respectively).\footnote{Simpler linear models such as logistic regression did not produce any usable results.} The feature set extracted from a document includes the document length in tokens, number of annotated arguments, average argument length, and for each argument type from our inventory its relative frequency in percent. Note that this feature-based approach is only applicable to documents annotated with arguments (which is the case for the MADON dataset) but cannot be used for inference on purely textual data, unlike the transformer models.

We employ grid search hyperparameter tuning for our experiments. We use dropout for each hidden layer with the rates of 0.1. and 0.4 for the first and the second layer, respectively. We also apply early stopping with the patience of 3 epochs to prevent the overfitting. Finally, we train the MLP model with the learning rate of 1e-3 and the batch size of 8. We also normalize the whole dataset based on the scaling factor of training dataset before the training, since the feature values are too sparse (see Section~\ref{subsec: madon_stats}).


\section{Resources} 
We report the resources we used for our experiments in Table \ref{tab:runtimes_all}.

\begin{table}
\centering
\caption{Training and inference runtimes for the best-performing Llama and ModernBERT configurations for argument presence detection (T1), argument type classification (T2), and holistic formalism classification (T3), trained on the full training set (189 decisions) and evaluated on the full test set (29 decisions). The last column reports the GPU accelerator used for each run.}
\label{tab:runtimes_all}
\begin{tabular}{llrrl}
\toprule
\textbf{Task} & \textbf{Model} & \textbf{Train} & \textbf{Test} & \textbf{GPU} \\
\midrule
\multirow{2}{*}{T1: Arg.\ Presence}
  & Llama (Base)        & 4h55m  & 43.0s  & A100 (80GB)      \\
  & ModernBERT (CPT)    & 1h19m  & 3.8s   & RTX A6000 (48GB) \\
\midrule
\multirow{2}{*}{T2: Arg.\ Type}
  & Llama (Base+Asy)    & 6h36m  & 300.0s & A100 (80GB)      \\
  & ModernBERT (CPT+Asy)& 2h12m  & 1.3s   & A100 (80GB)      \\
\midrule
\multirow{2}{*}{T3: Formalism}
  & Llama (CPT)         & 6h33m  & 13.0s  & A100 (80GB)      \\
  & ModernBERT (CPT)    & 32m    & 1.3s   & RTX A6000 (48GB) \\
\bottomrule
\end{tabular}
\end{table}

\end{appendices}

\end{document}